\documentclass[preprint,12pt]{elsarticle}

\usepackage{amssymb}

\journal{Neurocomputing}

\begin{document}

\begin{frontmatter}



\title{RUN:Residual U-Net for Computer-Aided Detection of Pulmonary Nodules without Candidate Selection}


\author[1]{Tian Lan}
\ead{lantian1029@uestc.edu.cn}

\author[1]{Yuanyuan Li\corref{cor1}}
\ead{lyy\_ann@126.com}

\author[1]{Jonah Kimani Murugi}
\ead{jkimani@seku.ac.ke}

\author[1]{Yi Ding}
\ead{yi.ding@uestc.edu.cn}

\author[1]{Zhiguang Qin}
\ead{qinzg@uestc.edu.cn}
\cortext[cor1]{Corresponding author}

\address[1]{School of Information and Software Engineering, University of Electronic Science and Technology of China, \\No.4, Section 2, North Jian she Road, Chengdu, Sichuan,China 610054 }


\begin{abstract}
The early detection and early diagnosis of lung cancer are crucial to improve the survival rate of lung cancer patients. Pulmonary nodules detection results have a significant impact on the later diagnosis. In this work, we propose a new network named RUN to complete nodule detection in a single step by bypassing the candidate selection. The system introduces the shortcut of the residual network to improve the traditional U-Net, thereby solving the disadvantage of poor results due to its lack of depth. Furthermore, we compare the experimental results with the traditional U-Net. We validate our method in LUng Nodule Analysis 2016 (LUNA16) Nodule Detection Challenge. We acquire a sensitivity of 90.90\% at 2 false positives per scan and therefore achieve better performance than the current state-of-the-art approaches.

\end{abstract}


\end{frontmatter}


\section{Introdution}
\label{}
Lung cancer has the highest morbidity and mortality in China. At present, surgery is still the only treatment that can cure lung cancer. Since it has no symptoms in its early stages, 70 -80\% of lung cancer patients are diagnosed when the cancer is already at an advanced stage, thereby losing the chance of undergoing a successful surgical treatment.  Nowadays, this situation is being improved with the development of artificial intelligence in the direction of medical image processing.
\par
The early manifestation of lung cancer in medical imaging is usually solitary pulmonary nodule (SPN). However, due to the large amount of information in the entire image, it is easy for human eyes to miss small nodules, and experts are also prone to misdiagnosis when they are tired. Therefore, computer-aided diagnosis (CAD) systems have been gradually developed. The early research of CAD of lung cancer is mainly the use of X-ray film. However, since X-ray imaging is based on the density of each detection site, it often misses tiny nodules and nodules hidden behind the heart and blood vessels, thereby leading to poor end results. With the continuous development and improvement of imaging technology, low-dose CT scanning gradually shows its superiority as it can even detect tumors as small as millimeters in size, and has become one of the most effective methods for detecting solitary pulmonary nodules in early stages of lung cancer. Accurate lung nodule detection is the key to CAD for lung cancer diagnosis and is usually divided into two main phases: nodule candidate detection and reduction of false positives. Presently, the method of nodule detection can be summarized as two types: 1) Traditional machine learning methods{\cite{Messay2010A,Filho20163D,Gon2016Hessian,Javaid2016A,Lan2018Detection}} : The region of interest (ROI) is extracted, then its characteristics are calculated, and finally classifiers are used for classification. Both the selection of features and classifiers will have a great impact on the final result. 2) Deep learning method \cite{Simonyan2014Very,Gruetzemacher2016Using,Ding2017Accurate,Li2016Pulmonary}: Build the network structure, train the model, and then use the trained model to classify the data. Since convolutional neural networks can autonomously learn features, the feature selection process of traditional methods is optimized. In this paper, we mainly study the classification of true and false nodules based on deep learning algorithms. At present, there are many deep learning models such as   CNN\cite{Krizhevsky2012ImageNet,Li2016Pulmonary}, DBN\cite{Mohamed2012Understanding,Li2015Classification}, RNN\cite{Liu2014A,Karpathy2014Deep}, GAN\cite{Goodfellow2014Generative,Radford2015Unsupervised} and so on.
In our experiment, we focused on other two kinds of models: U-Net \cite{Ronneberger2015U} and residual network\cite{He2015Deep}. 
\par 
U-Net, which was proposed in 2015\cite{Ronneberger2015U} by Olaf Ronneberger, Philipp Fischer, and Thomas Brox, won the International Symposium on Biomedical Imaging (ISBI) competition 2015. The entire network contains a total of 23 convolutional layers,including convolutions,max poolings,up-convolutions and a fully convolution. In general, it can be regarded as an encoder-decoder structure, the encoder gradually reduces the spatial dimension of the pooling layer, the decoder gradually repairs the details of the object and increase spatial dimensions.  There is a quick connection between the encoder and the decoder.  Residual network which was proposed by Kaiming He, Xiangyu Zhang in 2015 \cite{He2015Deep}, won the champion of ImageNet Large Scale Visual Recognition Competition (ILSVR) competition. It increases  the network depth without degrading by superimposing $y = x$ layers (called identity mappings) on a shallow network basis. The concept of a "shortcut" is proposed, which skips one or more layers and adds the input results to the bottom layer directly. The feature is extracted by adding multiple cascaded output results to the input, reducing the training parameters. Both the U-Net and residual network have a simple structure and faster training speed, but U-Net's depth is slightly insufficient, and residual network solves the problem of degeneration under extremely deep convolutional neural network effectively.
Therefore, we combine the two networks effectively and propose a new network called RUN. Compared with other methods, our biggest advantage can be summarized as: only one network is used to implement an end-to-end classification system directly without candidate nodules extraction.

\section{Methods}
\label{}
This paper presents an easy-to-implement lung nodule detection framework. On the one hand, the traditional method feature extraction process is optimized. On the other hand, we just use one network to  obtain the nodule detection results directly, which guarantees effectiveness and simplifies the detection process. The specific implementation of the system can be shown in Figure 1, which comprises only three stages of preprocessing(lung extraction), training of the model and classification, it is easy and effective.
\begin{figure}[ht]	
	\flushleft
	\includegraphics[scale=0.45]{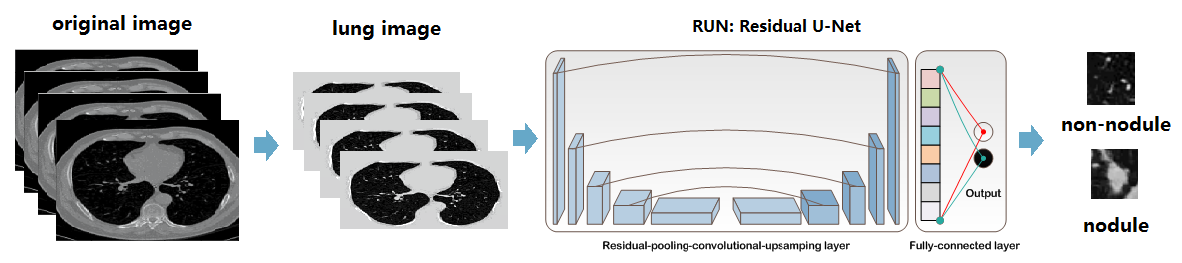}
	\caption{Pulmonary nodule detection framework. First, segmenting the lung parenchyma from the raw lung CT images and then using pre-processed images to train our network. In the end, nodules and non-nodules are classified by trained network.}
	\label{fig:label}
\end{figure}

\subsection{Preprocessing}
Preprocessing improves the overall system accuracy by enhancing image quality. At this stage, in order to remove the influence of background, the image is segmented using a threshold method and a morphology-related method after noise reduction, thereafter obtaining a refined lung image\cite{Gomathi2009Computer}.

\subsection{Improved network structure}
Although the original U-Net model is easy to train, the accuracy of the experimental results is affected to a certain extent due to the lack of depth. The Residual Attention Network\cite{Wang2017Residual} made the network model to reach the deep level by stacking the Attention Module. We therefore propose a method to introduce the main standpoint of the residual network into the U-Net. It not only deepens the depth of the network, but also guarantees the effectiveness of training. Being similar to Residual Attention Network, we stack the main components of the residual network: residual units, and each unit contains "shortcut connection" and "identity mapping." This deepens the network depth and ensures more detailed features simultaneously. The entire network is still in the form of a U-shaped structure, which is downsampling first and then upsampling, and the down-sampled feature map is merged with the corresponding up-sampled's. Finally, the classification result is obtained through the fully-connected layer.
\subsubsection{Residual unit}
For each residual unit, it can be expressed by the following formula:
\begin{equation}
x_{l+1}=H(x_{l})+F(x_{l};w_{l,k})
\end{equation}
$x_{l}$ and $x_{l+1}$ represent the input and output of the	$l_{th}$ residual unit, respectively.$w_{l,k}$ is the weight (and error) of the first residual unit, and k is the number of weighted layers contained in each residual unit ($k=2$). F represents a residual function, stacking two 3*3 convolutional layers. The function H is an identity mapping: $H(x_{l}) = x_{l}$. Rectified Linear Unit (ReLU) and  Batch Normalization(BN) as "pre-activation" of the weight layer. The specific residual unit design is shown in Figure 2, different from traditional structure\cite{He2016Identity}.
\begin{figure}[ht]	
	\centering
	\includegraphics[scale=0.92]{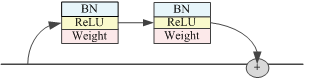}
	\caption{Residual Unit. It stacks two 3*3 convolutional layers, ReLU and BN as "preactivation" of every weight layer. }
	\label{fig:label}
\end{figure}

\subsubsection{Network structure}
Whole network structure is stacked into U-shaped and consists of downsampling and upsampling. In the downsampling process, the residual unit is introduced to deepen the network structure. In the upsampling process, in order to avoid the residual unit from transmitting more noise information, we only use a simple convolution operation.
\par
The network consists of a total of 10 residual unit layers, 4 max-pooling layers, 4 up-conv layers, 8 convolutional layers, and the finally 1 fully-connected layer. The structure of the network is shown in Figure 3. After each downsampling, size of the feature map is halved, and the number of feature maps is doubled; after each upsampling, the number of feature maps is halved and the size is doubled, and then merged with the corresponding feature maps in the downsampling process. To prevent over-fitting, we introduced the dropout operation during downsampling\cite{Srivastava2014Dropout}. At the same time, in order to speed up the convergence and further overcome the disadvantages of deep neural networks that are difficult to train, we used BN operations during the upsampling process\cite{Ioffe2015Batch}.
\begin{figure}[htp]	
	\centering
	\includegraphics[scale=0.98]{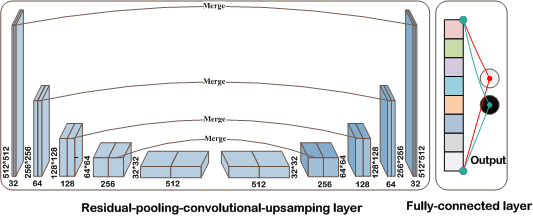}
	\caption{RUN architecture. Each box corresponds to a multi-channel  feature map.The number of channels is marked on bottom of the box and size of each feature map is marked next to the box. The light-colored box represents the residual unit, and the dark box represents the convolution.}
	\label{fig:label}
\end{figure}

\par
Like the U-Net network, downsampling can extract features, and upsampling can complete positioning; at the same time, the "shortcut" operation increases network depth and retains more original detail features. In contrast to the residual network which stacks residual unit directly, our approach uses different sized units stack into a U-shaped structure and it can avoid over-reliance on the performance of equipment.
\subsection{Loss function}
The loss function is used to measure the degree of inconsistency between the model's predicted value and the true value. In this experiment we just use the simple dice coefficient loss function, and it's definition is shown as follows: 
\begin{equation}
Loss=1-Dice 
\end{equation}
\par The calculation of the dice coefficient which is a similarity measure function is as follows:
\begin{equation}
Dice = \frac{2\left |X\bigcap Y  \right |}{\left | X \right | \bigcup \left | Y \right |}
\end{equation}
X represents the predicted value, and Y represents the true value. The $\left |X\bigcap Y  \right |$ represents the intersection of two sets, and the $\left | X \right | \bigcup \left | Y \right |$ represents the union. The more similar the two samples are, the closer the coefficient value is to 1.
Therefore, the larger the dice coefficient is, the smaller loss becomes and the better  robustness the model has.

\subsection{Optimization function}
The essence of most learning algorithms is to establish an optimization model, and optimize the objective function (or loss function) through the optimization method to train the best model. 
We used the adaptive moment estimation (Adam) optimization algorithm\cite{Kingma2014Adam} since each iteration has a certain range of learning rate after offset correction, making the parameters more stable. Essentially, the algorithm is a RMSprop with a momentum term, which dynamically adjusts the learning rate of each parameter using the gradient first moment estimation and the second moment estimation. Full calculations are listed as in Eq.(4) to Eq.(8): 
\begin{equation}
m_{t}=\beta _{1}\cdot m_{t-1}+(1-\beta_{1})\cdot g_{t}
\end{equation}
\begin{equation}
v_{t}=\beta _{2}\cdot m_{t-1}+(1-\beta_{2})\cdot g_{t}^{2}
\end{equation}
\begin{equation}
\widehat{m}_{t}=\frac{m_{t}}{1-\beta_{1}^{t}}
\end{equation}
\begin{equation}
\widehat{v}_{t}=\frac{v_{t}}{1-\beta_{2}^{t}} 
\end{equation}
\begin{equation}
\theta_{t}=\theta _{t-1}-\frac{\eta \cdot\widehat{m}_{t}}{\sqrt{\widehat{v}_{t}}+\epsilon } 
\end{equation}
where t represents the training iteration number, $ g$ is the gradient, $m$ and  $v$  signify the first moment estimate and the second moment estimate respectively, corresponding is that $\widehat{m}$ and $\widehat{v}$ denote the bias corrected first moment and the bias corrected second moment respectively, and Eq.(8) updates parameters finally. $\beta_{1},\beta_{2},\epsilon$ are adjustable parameters (general default is: $\beta_{1}=0.9,\beta_{2}=0.99,\epsilon=10^{-8}$), and $\eta$ represents the learning rate.

\section{Data and experiment}
\label{}

\subsection{Data processing}
In our experiments, all the data come from LUNA16. The data set is derived from Lung Image Database Consortium and Image Database
Resource Initiative (LIDC-IDRI) database, which includes 1018 research examples acquired from 1010 different patients. After picked scans with a slice thickness greater than 2.5mm, 888 CT scans are included in this challenge, and every scan contains annotations that were made by 4 experienced radiologists. This challenge consists all nodules ($>= 3mm$) accepted by at least 3 out of 4 radiologists. All the data is divided into 10 subsets, we use 9 of them for training and 1 for testing. To reduce the impact of ribs, scan’s intensity is clipped in range from -1200 up to 600 Housfield Unit and subsequently normalized to the range of [0, 1].
\subsection{Evalution Criterion }
 We use two evaluation criteria to analyze the performance of the network architecture  for the detection task. (1) We use Dice coefficient to evaluate our predicted results. When the Dice coefficient of the predicted value and the real value is greater than 50\%, we judge it as a hit. (2) The nodule area we predict contains the coordinates of the nodal centers labeled by the experts, it is also called a hit. Otherwise, it is determined there is no hit, what is a false positive.
\subsection{Experiment and results}

Instead of developing a whole pulmonary nodule detection
system, which usually integrates a candidate detector and a
FP reducer, our method completes the detection task only by one network. Due to the high computational cost, we use axial slices as inputs instead of the entire case to train\cite{Ding2017Accurate}. In whole training stage, each model's inputs are 512*512 images, with the per-pixel mean subtracted. The dropout($rate=0.2$) strategy is utilized in convolutional and fully connected layers to improve the generalization capability of each model.  The used batch size depends on the GPU memory and Adam is used to optimize model. When training the RUN, we use 60 epochs in total, the learning rate starts from 0.01, 0.001 after the epoch 10, and  0.0001 at the halfway of training. At testing stage, CT images are pre-processed the same way as we do in training stage. The networks are implemented in Python based on the deep learning framework Keras with Tensorflow backend using a GeForce  GTX 1080 Ti GPU. 
\par
Furthermore, in order to demonstrate the effectiveness of our RUN network structure, we adjust the corresponding parameter settings to train a U-Net and a dual-path residual U-Net. As Table 1 shows their network structures are similar to the RUN network structure we proposed. 
In U-Net, neither the downsampling nor the upsampling contain residual units. However, both the down-sampling and up-sampling processes use the residual unit as the basic component to stack  the dual-path residual U-Net network.  

\begin{table}[htp]
	\centering  
	\begin{tabular}{|c|c|c|} 
		\hline
		U-Net&Dual-Path Residual U-Net&RUN:Residual U-Net \\ \hline  
		conv-32&residual unit-32&residual unit-32 \\
		conv-32&residual unit-32&residual unit-32 \\  \hline
		\multicolumn{3}{|c|} {maxpooling} \\ \hline
		conv-64&residual unit-64&residual unit-64 \\
		conv-64&residual unit-64&residual unit-64 \\  \hline
		\multicolumn{3}{|c|} {maxpooling} \\ \hline
		conv-128&residual unit-128&residual unit-128 \\
		conv-128&residual unit-128&residual unit-128 \\  \hline
		\multicolumn{3}{|c|} {maxpooling}\\ \hline
		conv-256&residual unit-256&residual unit-256 \\
		conv-256&residual unit-256&residual unit-256 \\  \hline
		\multicolumn{3}{|c|} {maxpooling}\\ \hline
		conv-512&residual unit-512&residual unit-512 \\
		conv-512&residual unit-512&residual unit-512 \\  \hline	
		\multicolumn{3}{|c|} {up-conv+merge}\\ \hline
		conv-256&residual unit-256&conv-256 \\
		conv-256&residual unit-256&conv-256 \\  \hline
		\multicolumn{3}{|c|} {up-conv+merge}\\ \hline
		conv-128&residual unit-128&conv-128 \\
		conv-128&residual unit-128&conv-128 \\  \hline
		\multicolumn{3}{|c|} {up-conv+merge}\\ \hline
		conv-64&residual unit-64&conv-64 \\
		conv-64&residual unit-64&conv-64 \\  \hline
		\multicolumn{3}{|c|} {up-conv+merge}\\ \hline
		conv-32&residual unit-32&conv-32 \\
		conv-32&residual unit-32&conv-32 \\  \hline
		\multicolumn{3}{|c|} {fully-connected conv}\\ \hline
	\end{tabular}
	\caption{Comparison of network structure of three different networks.}
\end{table}
\begin{table}[htbp]
	\centering  
	\begin{tabular}{|c|c|c|} 
		\hline
		Network Structure&Training Parameters&Dice Coefficient \\ \hline  
		RUN:Residual U-Net  &8,812,837&71.9\% \\ \hline
		Traditional U-Net&8,631,841&66\%\\ \hline        
		Dual-Path Residual U-Net& 8,993,157 & 63.05\%\\ \hline        
		
	\end{tabular}
	\caption{Comparison of experiment results with three different networks.}
\end{table}
\par
The comparison of experiment results is shown in Table 2 and Figure 4. From Table 2, we can see that because introducing residual unit can deepen the depth of the network, the RUN network has more training parameters than the U-Net but less than the dual-path residual U-Net. And besides the RUN network segmentation effect is significantly higher than that of the U-Net, and its effect is improved by about 6\%, but the dual-path residual U-Net results in more erroneous segmentation. 
\begin{figure}[htp]	
	\centering
	\includegraphics[scale=0.625]{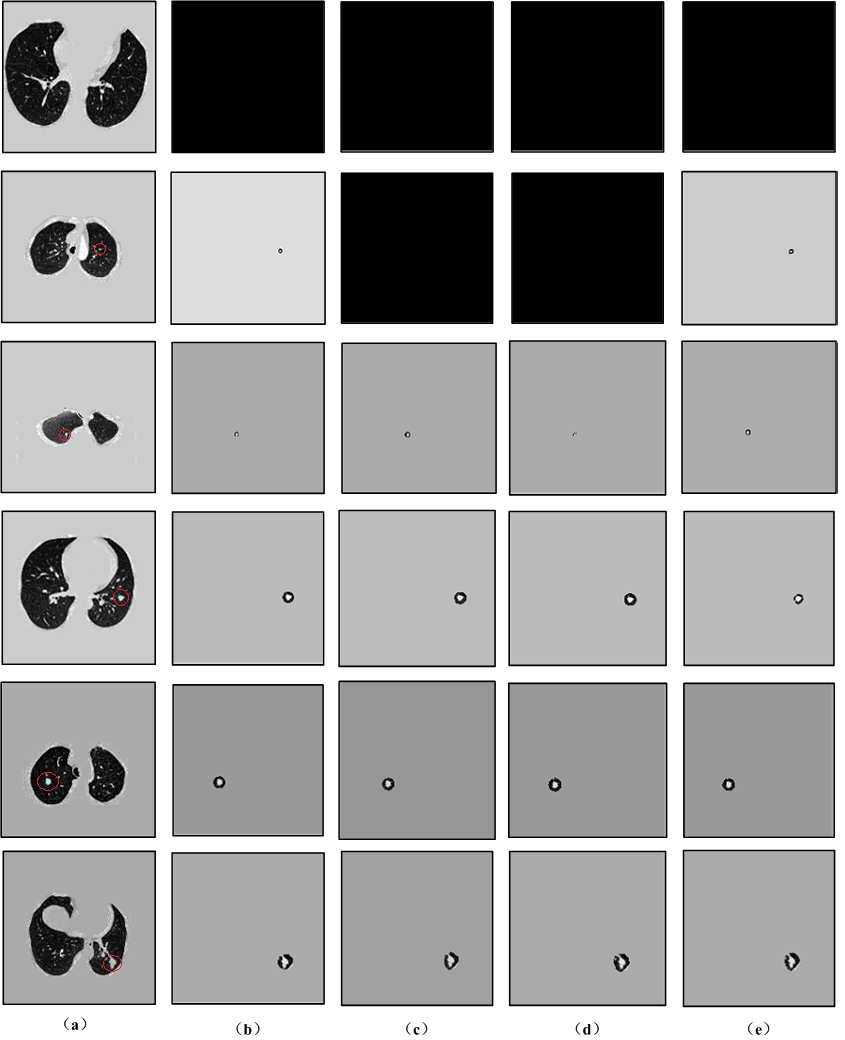}
	\caption{The results of the segmentation of the same CT image by three different networks. All black indicates there are no nodules. (a) Pre-processed CT image, red circles circled with nodule. (b) Nodule region in CT image. (c) Nodule segmented after training through dual-path residual U-Net. (d) Nodule segmented after training through U-Net. (e) Nodule segmented after training through RUN.}
	\label{fig:label}
\end{figure}

In figure 4, we show  results of the segmentation of the same CT image by three different networks. CT image of each row indicates small nodule, smaller nodule, middle nodule, larger nodule and large nodule, respectively. From these segmentation results, it can be clearly seen that, when the segmentation object is small nodules, RUN is obviously  better than the other two methods.

\begin{figure}[ht]	
	\centering
	\includegraphics[scale=0.75]{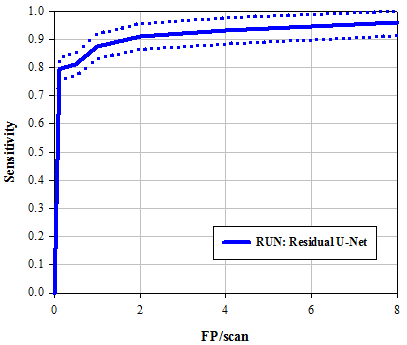}
	\caption{Free-Response Receiver Operating Characteristic(FROC) curves. The dash lines
		are lower bound and upper bound FROC for 95\% confidence interval.}
	\label{fig:label}
\end{figure}

\par
 Currently, the FPs/scan rates between 1 and 4 are mostly preferred in clinical practice\cite{Ginneken2010Comparing}.  As Figure 5 shows, our method yields a sensitivity of 79.05\% at 0.11 FPs/scan and 90.90\% at 2 FPs/scan. Therefore, it can be noted that the results from our experiments can satisfy clinical usage.  In addition, in order to show the performance of the proposed method, we compare it with the other state-of-the-art methods designed for lung nodule detection. The result is shown in Table 3.
\begin{table}[ht]
	\centering  
	\begin{tabular}{|l|c|c|c|} 
		\hline
		Lung nodule detection sys. &Cases&Sensitivity(\%)&FP(per case) \\ \hline  
		Messay et al.\cite{Messay2010A}&84&82.66&3\\  \hline        
		Bergtholdt et al.\cite{Bergtholdt2016Pulmonary} &243&85.9 &2.5 \\   \hline
		Li et al.\cite{Li2016Pulmonary}&1010&87.1 &4.622 \\   \hline
		Golan et al.\cite{Golan2016Lung} &1018&71.2 &10 \\   \hline
		Huang et al.\cite{Huang2017Lung}&99 &90 &5 \\   \hline
		Setio et al.\cite{Setio2016Pulmonary}  &888&90.1 &4 \\   \hline
		Dou et al.\cite{Dou2017Automated}  &888&90.6 &2 \\   \hline
		The proposed method &888&90.9&2\\ \hline
	\end{tabular}
	\caption{Performance comparison with other lung nodule detection methods}
\end{table}
\section{Discussion}
We introduce the "shortcut" of the residual network to improve the traditional U-Net to get a RUN for pulmonary nodule detection. Due to pulmonary nodules vary
greatly in size (diameter range from 3mm to 30 mm), many existing successful detection and diagnosis systems employ a multi-scale architecture over the years\cite{Setio2016Pulmonary,Gori2007A,Wei2015Multi,Dou2017Multi}. However, these methods need different sizes of nodule patches as input and to adjust the size of the receptive field according to the nodule size, the setting of the receptive field is very important to the results. Therefore, during the training of our experiment, the entire slice is used as an input to the network instead of the patch. 
We also verify the influence of spatial information of different size nodules on detection task by studying the effect of single slice and multi-slice training on the experimental results, and we find that the presence of very small nodules may result in the false learning of multi-slice training and lead to more FP, but in general, more spatial information is taken into account, which can help reduce misdiagnosis effectively. Therefore, if we can choose the correct number of slices for training according to the size of the nodules, the results must be improved significantly.

\section{Conclusion}
\label{}
In this paper, we present a network called RUN for
computer-aided detection of pulmonary nodules skipped the stage of candidate selection from volumetric CT scans. We prove this network has a good learning performance with complex and variable lung nodules via LUNA16. In principle, the proposed framework is generic and can easily be extended to other target detection tasks in medical images. Further investigations include the evaluation of more clinical data and the investigation of more methods to achieve better experimental results for clinical use, such as the three classification problems of pulmonary nodules.









\end{document}